\documentclass[runningheads]{llncs}
\usepackage[T1]{fontenc}
\usepackage{graphicx,verbatim}
\usepackage{multirow}
\usepackage{booktabs}
\usepackage[table]{xcolor}
\usepackage{amssymb}
\usepackage{pifont}
\usepackage{amsmath}
\usepackage[colorlinks=true,
            linkcolor=blue,
            anchorcolor=blue,
            citecolor=blue,
            urlcolor=blue,
            ]{hyperref}
\usepackage[table]{xcolor}

\newcommand{\best}[1]{\cellcolor{red!12}{#1}}
\newcommand{\second}[1]{\cellcolor{blue!10}#1}
            
\newcommand{\kiconimg}{\IfFileExists{imgs/icon_kidney.pdf}{\includegraphics[height=0.82em]{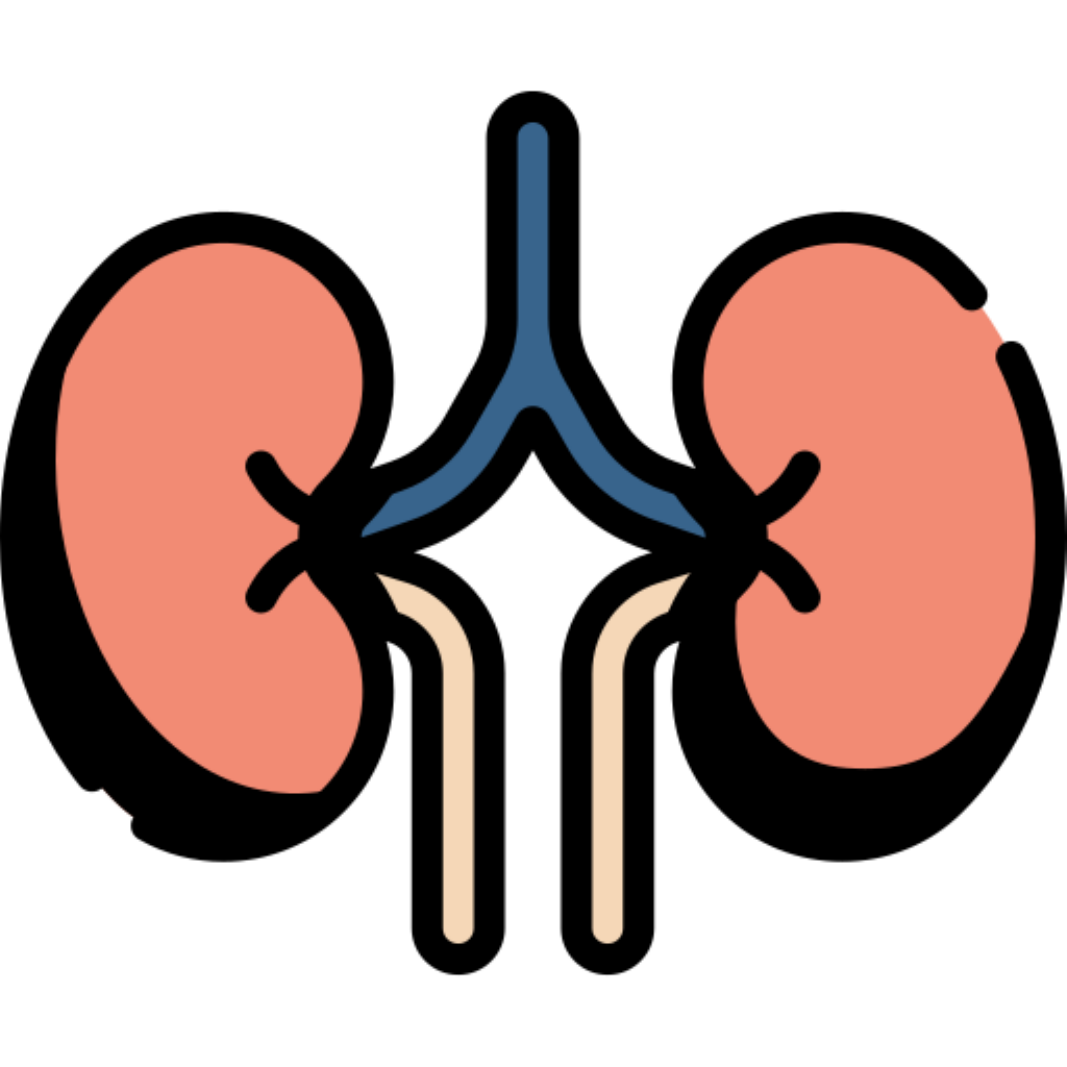}}{\includegraphics[height=0.82em]{imgs/icon_kidney.pdf}}}
\newcommand{\liconimg}{\IfFileExists{imgs/icon_liver.pdf}{\includegraphics[height=0.82em]{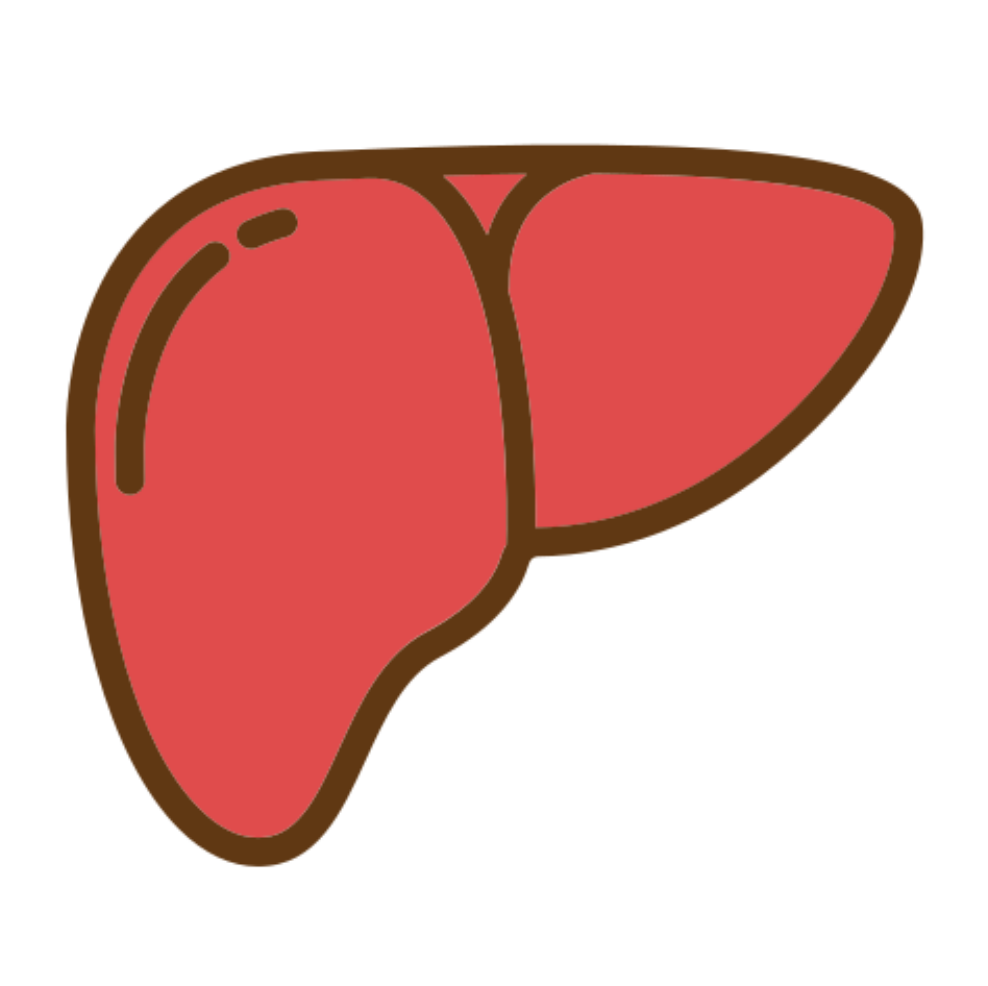}}{\includegraphics[height=0.82em]{imgs/icon_liver.pdf}}}
\newcommand{\piconimg}{\IfFileExists{imgs/icon_pancreas.pdf}{\includegraphics[height=0.82em]{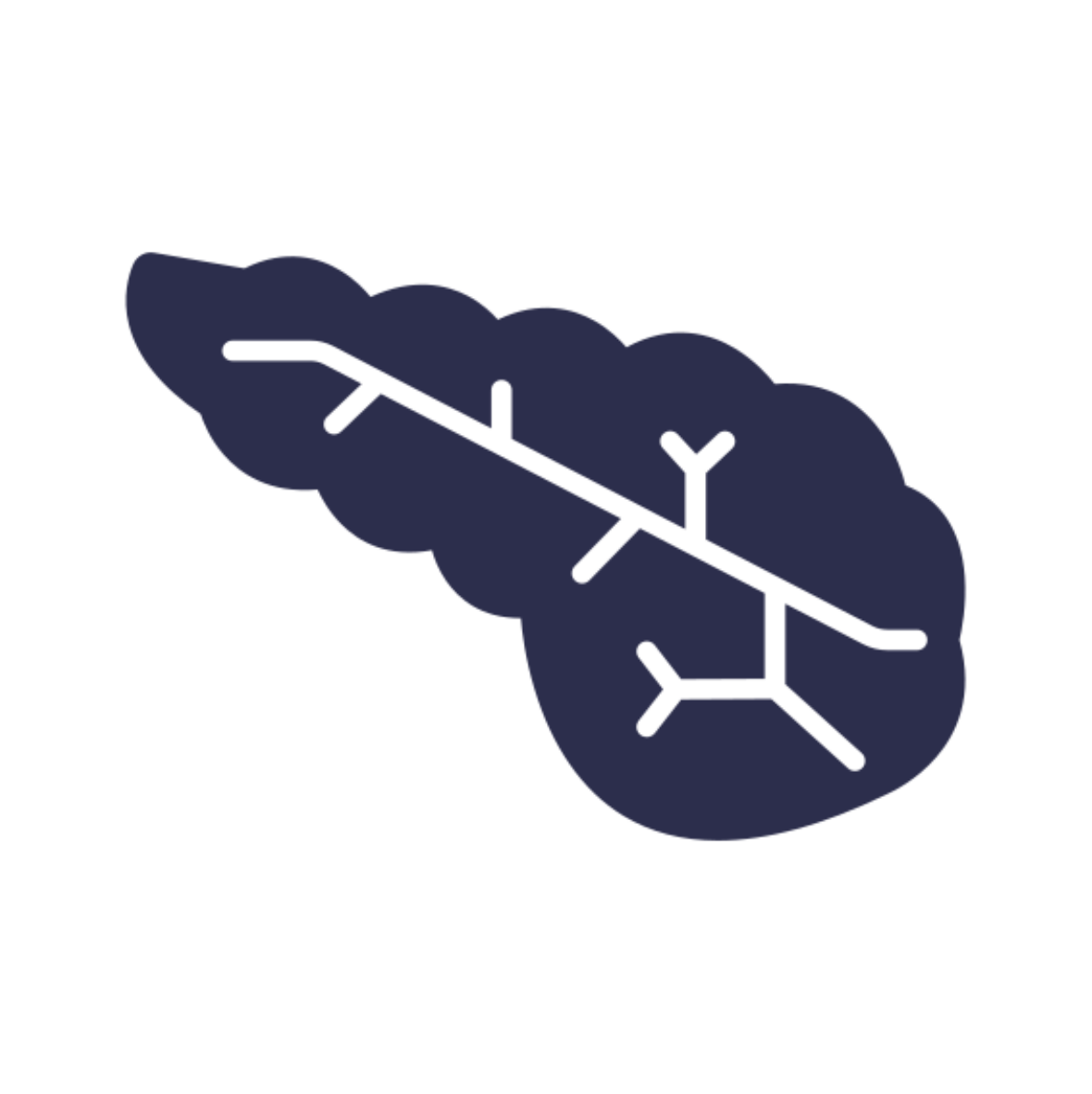}}{\includegraphics[height=0.82em]{imgs/icon_pancreas.pdf}}}
\newcommand{\kicon}{K\kern0.08em\smash{\raisebox{-0.12ex}{\kiconimg}}}
\newcommand{\licon}{L\kern0.08em\smash{\raisebox{-0.12ex}{\liconimg}}}
\newcommand{\picon}{P\kern0.08em\smash{\raisebox{-0.12ex}{\piconimg}}}
\begin{document}
%
\title{GLeVE: Graph-Guided Lesion Grounding with Proposal Verification in 3D CT}
\titlerunning{GLeVE: Graph-Guided Lesion Grounding with Proposal Verification}
%
\author{%
  Shuo Jiang\inst{1}$^\dagger$ \and 
  Yuhao Hong\inst{1}$^\dagger$ \and 
  Chunbo Jiang\inst{1} \and 
  Weihong Chen\inst{1} \\ 
  Huangwei Chen\inst{2} \and 
  Shenghao Zhu\inst{1} \and 
  Beining Wu\inst{1} \and 
  Mingxuan Liu\inst{3} \and 
  Zhu Zhu\inst{4} \\ 
  Feiwei Qin\inst{1}\thanks{Corresponding authors.\quad $\dagger$ Co-first authors.} \and 
  Min Tan\inst{1} \and 
  Yifei Chen\inst{3}\protect\footnotemark[1] 
}

\authorrunning{S. Jiang et al.}

\institute{%
  Zhejiang Key Laboratory of Space Information Sensing and Transmission, Hangzhou Dianzi University, Hangzhou, China
  \and
  Zhejiang University, Hangzhou, China
  \and
  Tsinghua University, Beijing, China
  \and
  Children's Hospital, Zhejiang University School of Medicine, Hangzhou, China
  \\
  \email{\{jiangshuo, qinfeiwei\}@hdu.edu.cn, justlfc03@gmail.com}
}
\maketitle              
%



\begin{abstract}
Grounding radiology report descriptions to 3D CT volumes is essential for verifiable clinical interpretation, yet remains challenging due to the semantic-spatial gap between free-text narratives and volumetric anatomy. Existing report-assisted and vision-language grounding methods typically rely on phrase-level alignment or dense pixel supervision, resulting in limited lesion-wise correspondence and suboptimal localization accuracy. We propose GLeVE, a graph-guided lesion grounding framework with anatomical prior verification and octree-based autoregressive refinement. GLeVE treats each lesion description as an atomic semantic unit and encodes organ attribution, attributes, and inter-lesion relations through relation-aware graph reasoning to produce discriminative lesion-wise queries. Anatomy-aware proposal generation with region-level verification enforces one-to-one text-lesion alignment, while hierarchical octree refinement progressively improves boundary delineation. Experiments on AbdomenAtlas 3.0 demonstrate consistent gains over classical multimodal foundation models and report-supervised baselines in both segmentation accuracy and lesion-level localization. Our source code is available at \href{https://github.com/JSLiam94/GLeVE/}{https://github.com/JSLiam94/GLeVE}.
\end{abstract}

\section{Introduction}

Radiology reports provide structured natural-language descriptions of lesions in 3D Computed Tomography (CT), covering organ attribution, anatomical subregions, and key imaging characteristics such as Hounsfield Units (HU), and thus serve as a central interface between imaging evidence and clinical decision-making~\cite{kulsoom2025natural,busch2025large,bai2026exact}. However, in routine clinical workflows, physicians must repeatedly revisit the original CT and search slice by slice around the reported locations to verify critical findings, making lesion localization highly experience-dependent and inefficient~\cite{li2025automatic,dong2025keyword,bai2025chest}. This burden is further amplified in multi-organ/multi-lesion scenarios: reports grow substantially longer, salient lesion descriptions are diluted by abundant routine statements, and clinicians are hindered in rapidly retrieving key information from lengthy text and stably mapping it to the corresponding image locations, thereby limiting the practical utility of reports for precise decision-making and verifiable interpretation~\cite{alabed2026large,tejani2024checklist,chen2025mmlnb}.

Existing general-purpose segmentation models (e.g., SwinUNETR~\cite{2022swin} and MedSAM3~\cite{2025medsam3}) produce global CT masks, hindering interpretable lesion-by-lesion report alignment and requiring dense pixel-level supervision~\cite{chen2024scunet++}. R-Super~\cite{bassi2025learning} uses reports as auxiliary supervision to reduce annotation reliance, but without explicit inference-time matching, it struggles to enforce one-to-one ``text-lesion'' correspondence. Studies on medical text grounding aim to align report descriptions into the global image coordinate system for explainable localization: Ichinose \emph{et al.}~\cite{Ichinose2023miccai} leverage anatomical-structure segmentation and structured report parsing to improve abnormality localization, but their reliance on cross-modal attention alignment still limits spatial precision; N{\"u}tzel \emph{et al.}~\cite{nutzel2025generate} employ latent diffusion models for bimodal bias fusion, and Vilouras~\cite{ZeroShot2025} performs feature selection via cross-attention maps for medical phrase grounding. Nevertheless, these methods mainly perform phrase-level alignment and cannot reliably treat a full lesion description as a coherent semantic unit for precise localization. Furthermore, Multimodal Large Language Model-based methods introduce parameter-efficient fine-tuning and uncertainty-aware multi-hypothesis mechanisms~\cite{he2025parameter,zou2025uncertainty} to enhance robustness; Yang \emph{et al.}~\cite{yang2026multimodal} propose a contrastive learning framework with multi-level semantic alignment, but it remains highly sensitive to fine-grained report quality and shows limited cross-scenario generalization. Overall, while prior work has advanced bounding-box-level localization, substantial gaps remain in achieving reliable, pixel-accurate localization of multiple low-contrast small lesions, falling short of core clinical requirements.

\begin{figure}[t!]
\includegraphics[width=\textwidth]{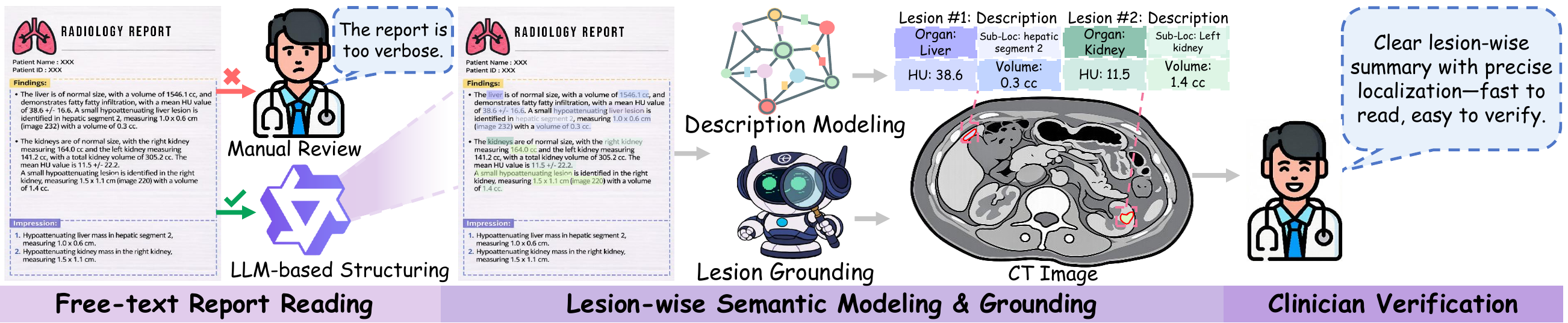}
\caption{Overview of lesion-wise report grounding, illustrating structured semantic modeling and precise CT localization for efficient clinical verification.} 
\label{Fans}
\end{figure}

To address these issues, we propose \textbf{G}raph-Guided \textbf{Le}sion Grounding with Proposal \textbf{Ve}rification framework (\textbf{GLeVE}). \textbf{GLeVE} encodes composite semantics in lesion descriptions, including organ attribution, size, and HU, through graph-structured representation learning, and incorporates organ-segmentation priors to construct geometry-aware anatomical embeddings. A multi-candidate verification mechanism is subsequently introduced to suppress boundary noise and spurious responses. This is followed by an octree-based autoregressive refinement strategy that enhances pixel-level localization sensitivity for small lesions. As illustrated in Fig.~\ref{Fans}, by optimizing semantic consistency together with spatial precision in complex anatomy, \textbf{GLeVE} delivers verifiable pixel-level lesion localization with one-to-one alignment between key report descriptions and imaging evidence, yielding traceable, text-grounded results for reliable clinical interpretation and quantitative evaluation.

\section{Proposed Method}

\begin{figure}[t!]
\includegraphics[width=\textwidth]{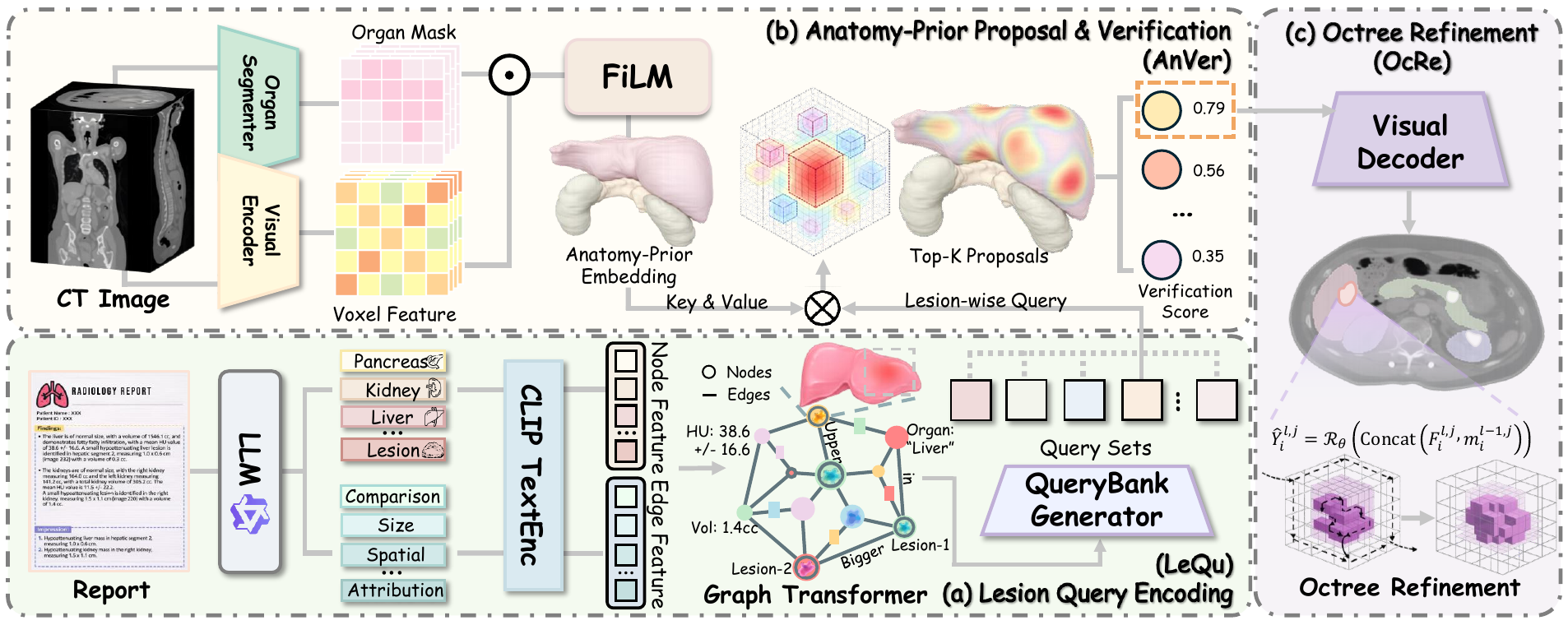}
\caption{Architecture of \textbf{GLeVE}, including (a) lesion semantic graph modeling, (b) anatomy-aware proposal verification, and (c) octree-based autoregressive refinement.}
\label{model}
\end{figure}

\subsection{Model Architecture}
As illustrated in Fig.~\ref{model}, \textbf{GLeVE} comprises three modules: (i) a graph-structured semantic query generator that converts each lesion description into lesion-wise queries while preserving organ attribution, attributes, and inter-lesion relations; (ii) anatomy-prior feature modulation with voxel-level similarity to generate high-recall candidates, followed by region-level verification for one-to-one text-lesion alignment; and (iii) octree-based autoregressive refinement within selected candidates for high-resolution segmentation, yielding interpretable masks consistent with global spatial coordinates.

\subsection{Lesion Semantic Modeling and Querying (LeQu)}

For lesion-level semantic completeness and precise cross-modal matching, we use each full lesion description as the atomic reasoning unit, encode its attribute constraints and spatial relations in an explicit graph, and generate corresponding lesion-wise queries for stable, interpretable localization in multi-lesion settings.

\textbf{Structured Report Parsing and Lesion Unit Construction.}
Given a report $R$, we employ a frozen Qwen3-8B~\cite{yang2025qwen3} to extract a lesion set $\{r_i\}$ with structured fields (organ attribution ${o(i)}$, sub-location, size, and HU), and incorporate organ-level statistics to explicitly model lesion-background contrast.

\textbf{Lesion Semantic Graph Construction.}
The extracted fields are organized into a semantic graph $G_i=(\mathcal{V}_i,\mathcal{E}_i)$ comprising lesion, anatomical, and attribute nodes. Edges encode lesion-organ attribution, lesion-attribute associations, and intra-organ relations, enabling cross-lesion comparison to disambiguate multi-lesion cases and enhance lesion-level localization queries.

\textbf{Node Initialization and Relation Encoding.}
For each node $v\in\mathcal{V}_i$, we initialize its embedding $h_v^{0}\in\mathbb{R}^{d}$ with a CLIP text encoder; for each edge $(u\rightarrow v)\in\mathcal{E}_i$, we assign a learnable embedding $r_{uv}$ to encode relation types.

\textbf{Relation-Aware Graph Transformer Reasoning.}
We adopt a relation-type-aware attention mechanism for message passing, where relation embeddings are injected into the graph reasoning process as attention biases:
\begin{equation}
\begin{gathered}
\alpha_{uv}^{\ell} = 
\mathrm{softmax}_{u\in\mathcal{N}(v)}
\left(
\frac{
\left(W_q h_v^{\ell-1}\right)^{\top}
\left(W_k h_u^{\ell-1} + W_r r_{uv}\right)
}{\sqrt{d}}
\right),\\
h_v^{\ell} = 
\mathrm{FFN}\left(h_v^{\ell-1}\right) 
+ 
\sum_{u \in \mathcal{N}(v)} 
\alpha_{uv}^{\ell} \, W_v h_u^{\ell-1},
\end{gathered}
\end{equation}
where $h_v^{\ell}\in\mathbb{R}^{d}$ is the $\ell$-th layer embedding of node $v$, $\mathcal{N}(v)$ denotes its neighborhood, $W_q,W_k,W_v\in\mathbb{R}^{d\times d}$ are the query/key/value projections, and $W_r\in\mathbb{R}^{d\times d_r}$ projects relation embeddings; $d$ is the feature dimension, and $\mathrm{FFN}(\cdot)$ is a residual Feed-Forward Network. This facilitates relation-aware message passing and enhances lesion-level semantic discrimination in multi-lesion settings.

\textbf{Lesion-Level Query Generation.}
After graph reasoning, the final representation of the lesion node $h_{v_i^{\mathrm{les}}}^{L}$ is treated as the semantic summary vector $z_i$ for the $i$-th lesion. A QueryBank Generator ($\Psi_{\mathrm{QB}}$) then maps $z_i$ to a set of lesion queries,
$
\left\{ q_i^{(m)} \right\}_{m=1}^{M} 
= 
\Psi_{\mathrm{QB}}(z_i),
$
where $M$ denotes the number of queries associated with each lesion.

\subsection{Anatomy-Prior Lesion Proposal and Verification (AnVer)}

To incorporate anatomical constraints, we combine organ embeddings with feature modulation to render voxel features organ-aware prior to matching, thereby enforcing consistency in candidate generation and region-level verification.

\textbf{Anatomical Prior Soft Modulation.}
The visual encoder produces features $F$.
An organ segmenter provides binary masks $\{M_k\}_{k=1}^{K_o}$, where $K_o$ denotes the total number of organ categories, and we Gaussian-smooth them to obtain soft maps $\{\widetilde{M}_k\}_{k=1}^{K_o}$.
For each organ, we learn an embedding $e_k$ and form a voxel-wise anatomy token that conditions Feature-wise Linear Modulation (FiLM):
\begin{equation}
\begin{gathered}
\hat{E}(x)=\phi_o\!\left(\sum_{k=1}^{K_o}\widetilde{M}_k(x)\,e_k\right)\in\mathbb{R}^{C},\\
(\gamma(x),\beta(x))=\psi\!\left(\hat{E}(x)\right),\qquad
\widetilde{F}(x)=\gamma(x)\odot F(x)+\beta(x),
\end{gathered}
\end{equation}
where $C$ denotes the channel dimension of the visual feature map, $\phi_o$ is a linear projection from the organ-embedding space to $\mathbb{R}^{C}$, and $\psi$ is a lightweight MLP that predicts the FiLM parameters $\gamma(x),\beta(x)\in\mathbb{R}^{C}$.
This channel-wise modulation enhances organ-consistent responses while
suppressing irrelevant regions, injecting anatomical awareness without
altering global spatial structure.

\textbf{Voxel-Level Similarity Response and Candidate Generation.}
For the $i$-th lesion, its query set $\{q_i^{(m)}\}_{m=1}^{M}$ is matched against the organ-aware feature map $\widetilde{F}$ via voxel-level cosine similarity, yielding a response map $S_i$:
\begin{equation}
S_i(x)
=
\max_{m}
\frac{
\left\langle
q_i^{(m)},\, \widetilde{F}(x)
\right\rangle
}{
\left\| q_i^{(m)} \right\|
\left\| \widetilde{F}(x) \right\|
}.
\end{equation}

Based on $S_i$, adaptive thresholding and 3D connected-component analysis are performed to select the Top-$K_p$ high-response components, forming a high-recall candidate set 
$
\mathcal{P}_i = \{P_i^{(k)}\}_{k=1}^{K_p}.
$
Each candidate $P_i^{(k)}$ retains its spatial indices in the global CT coordinate system and serves as a localization anchor for subsequent verification and segmentation refinement.

\textbf{Region Consistency Verification.}
Since voxel-level similarity peaks are prone to noise, we verify candidates at the region level. For lesion description $i$ with candidates $\{P_i^{(k)}\}_{k=1}^{K_p}$, we crop and pool the feature map $\widetilde{F}$ to obtain region features $v_i^{(k)}$, and construct evidence vectors $\xi_i^{(k)}$ encoding anatomical coverage and report-attribute consistency. Specifically, $\xi_i^{(k)}$ includes the overlap ratio with $\widetilde{M}_{o(i)}$ and region statistics from $P_i^{(k)}$, i.e., volume $V_i^{(k)}$ and mean HU $\mu_i^{(k)}$, which are compared with reported volume and HU to derive matching metrics:
\begin{equation}
s_i^{(k)}
=
\sigma
\left(
W_s
\left[
v_i^{(k)} ; z_i
\right]
\right)
+
\boldsymbol{\eta}^{\top} \xi_i^{(k)},
\qquad
P_i^{\ast}
=
\arg\max_{k}
\, s_i^{(k)},
\end{equation}
where $z_i$ denotes the lesion-level semantic vector, $\sigma(W_s[\cdot])$ produces a scalar semantic matching score, and $W_s$ and $\boldsymbol{\eta}$ are learnable parameters.

To promote a single dominant match per description, we minimize the entropy of the temperature-scaled candidate distribution:
\begin{equation}
\mathcal{L}_{\mathrm{uni}}
=
-
\sum_{i=1}^{N}\sum_{k=1}^{K_p}
p_i^{(k)}\log\!\left(p_i^{(k)}+\epsilon\right),\qquad
p_i^{(k)}=\frac{\exp\!\left(s_i^{(k)}/\tau\right)}{\sum_{j=1}^{K_p}\exp\!\left(s_i^{(j)}/\tau\right)},
\end{equation}
where $p_i^{(k)}$ is the temperature-scaled soft assignment, $\tau$ controls distribution sharpness, and $\epsilon$ is a small constant for numerical stability.

\subsection{Octree Autoregressive Lesion Refinement (OcRe)}

We propose an octree-driven autoregressive refinement framework for coarse-to-fine, hierarchical optimization of lesion masks, improving boundary delineation for small lesions with blurred details while remaining computationally efficient.

\textbf{Octree Construction and Hierarchical Refinement.}
From the selected candidate $P_i^{\ast}$, we build an octree $\mathcal{T}_i$ with root cube $\Omega_i$ enclosing $P_i^{\ast}$; recursive subdivision yields multi-level blocks $\Omega_i^{\ell,j}$ across spatial scales, progressively narrowing the region of interest in a coarse-to-fine, top-down refinement hierarchy.

\textbf{Autoregressive Mechanism and Conditional Modulation.}
At each refinement level, we employ a conditional autoregressive strategy, where the prediction from the previous level serves as an explicit structural prior for progressive segmentation refinement. 
The initial coarse mask $\hat{Y}_i^{0}$ is initialized from the selected candidate $P_i^{\ast}$.
For each octree node $\Omega_i^{\ell,j}$, we crop local organ-aware features,
$
F_i^{\ell,j} = \mathrm{Crop}(\widetilde{F}, \Omega_i^{\ell,j}),
$
and use the upsampled prediction from the preceding level as a conditional input,
$
m_i^{\ell-1,j} = \mathrm{Crop}\!\left(\mathrm{Up}(\hat{Y}_i^{\ell-1}), \Omega_i^{\ell,j}\right).
$
The concatenated representation is then fed into a parameter-shared refinement decoder:
\begin{equation}
\hat{Y}_i^{\ell,j}
=
R_{\theta}\!\left(
\mathrm{Concat}(F_i^{\ell,j},\, m_i^{\ell-1,j})
\right),
\label{eq:octree_refine}
\end{equation}
where $R_{\theta}$ is a lightweight 3D residual refinement head shared across levels and nodes. The node-level predictions $\{\hat{Y}_i^{\ell,j}\}_{j}$ are merged into $\hat{Y}_i^{\ell}$ for coarse-to-fine boundary refinement, especially around small and low-contrast lesions.

\textbf{Loss Under Missing Mask Annotations.}
Since many clinical samples lack pixel-level masks, we optimize the model on unlabeled data via report-driven weak supervision: for lesion $i$, the prediction yields $V_i$ and $\mu_i$ to match $V_i^{r}$ and $\mu_i^{r}$, while $\mathcal{L}_{\mathrm{sep}}$ penalizes inter-lesion overlap; the weak loss is defined as:
\begin{equation}
\mathcal{L}_{\mathrm{weak}}
=
\lambda_{\mathrm{uni}}\mathcal{L}_{\mathrm{uni}}
+
\lambda_{\mathrm{con}}\mathcal{L}_{\mathrm{con}}
+
\lambda_{\mathrm{sep}}\mathcal{L}_{\mathrm{sep}},
\end{equation}
where $\mathcal{L}_{\mathrm{con}}=\mathcal{L}_{\mathrm{attr}}+\mathcal{L}_{\mathrm{org}}$,
$\mathcal{L}_{\mathrm{attr}}=\sum_{a\in\{V,\mu\}}\sum_{i=1}^{N}
\big(\frac{|a_i-a_i^{r}|}{|a_i^{r}|+\epsilon}\big)^{2}$,
$\mathcal{L}_{\mathrm{org}}$ penalizes predicted lesion mass outside the binary target-organ mask $M_{o(i)}$ and $\lambda_{\mathrm{uni}},\lambda_{\mathrm{con}},\lambda_{\mathrm{sep}}$ weight unimodality, consistency, and separation.

For samples with Ground Truth (GT) masks, a combined Dice and cross-entropy loss $\mathcal{L}_{\mathrm{seg}}$ is employed. Let $\mathcal{D}_m$ denote the subset of samples with pixel-level labels. For each sample $x$, the total loss is defined as:
\begin{equation}
\begin{gathered}
\mathcal{L}_{\mathrm{total}}
=
\delta\,\lambda_{\mathrm{seg}}\,\mathcal{L}_{\mathrm{seg}}
+
\lambda_{\mathrm{weak}}\,\mathcal{L}_{\mathrm{weak}},\quad\text{s.t.}\quad
\delta = \mathbb{I}[x \in \mathcal{D}_m].
\end{gathered}
\end{equation}

\section{Experiments}

\subsection{Dataset and Implementation}
\textbf{Dataset.}
We conduct experiments on AbdomenAtlas 3.0~\cite{bassi2025radgpt}, comprising 9{,}262 cases across 138 institutions. Each case includes a report reviewed by 12 board-certified radiologists and voxel-aligned tumor masks for abdominal tumors, including kidney (K, 2{,}122), liver (L, 1{,}472), and pancreas (P, 361). We randomly split the cases into training, evaluation, and test sets at an 8:1:1 ratio.

\textbf{Implementation Details.}
All experiments are implemented in PyTorch~2.4 and conducted on an NVIDIA H100 GPU. We use MedFormer~\cite{medformer2023} as the visual encoder-decoder backbone and adopt TotalSegmentator~\cite{total} for organ segmentation. We set the batch size to 1 and train for 100 epochs with the random seed of 2026. Model parameters are optimized using AdamW~\cite{Loshchilov2017DecoupledWD} with an initial learning rate of $1 \times 10^{-4}$ and cosine annealing for learning-rate decay.

\textbf{Evaluation metrics.}
For evaluation, we report Dice and HD95 and additionally adopt Lesion Recall (LR) and Lesion Localization Score (LLS). We extract lesion instances via connected-component analysis on prediction and GT masks, declare a match when $\mathrm{Dice}\ge 0.1$, and use the Hungarian algorithm for optimal assignment, yielding $N_{\mathrm{matched}}$ and $N_{\mathrm{gt}}$. LLS explicitly penalizes centroid displacement, yielding greater sensitivity to lesion-level localization errors.

\begin{equation}
\mathrm{LR}
=
\frac{N_{\mathrm{matched}}}{N_{\mathrm{gt}}},\qquad
\mathrm{LLS}
=
\frac{1}{N_{\mathrm{gt}}}
\sum_{i=1}^{N_{\mathrm{gt}}}
\mathrm{Dice}\!\left(i,\pi(i)\right)\,
\exp\!\left(-\frac{d_i}{d_0}\right),
\end{equation}
where $d_i$ is the centroid distance for the $i$-th match, $d_0=20~\mathrm{mm}$ is the decay scale, and $\mathrm{Dice}(i,\pi(i))=0$ for unmatched GT lesions.

\begin{table}[t!]
\caption{Quantitative comparison under varying mask ratios across \textbf{GLeVE} and other methods. \colorbox{red!12}{Red} and \colorbox{blue!10}{blue} cells denote the best and second-best results, respectively.
}
\label{tab-com}
\centering
\fontsize{8pt}{10pt}\selectfont
\setlength{\tabcolsep}{0.65pt}
{
\renewcommand{\arraystretch}{0.88}
\begin{tabular*}{\linewidth}{@{\extracolsep{\fill}}l|ccc|ccc|c|ccc}
\toprule
\multirow{2.3}{*}{\textbf{Method}} 
& \multicolumn{3}{>{\columncolor{green!6}}c|}{\textbf{Dice$_{\smash{\scalebox{0.80}[1]{$\scriptscriptstyle std$}}}$ (\%)$\uparrow$}} 
& \multicolumn{3}{>{\columncolor{orange!5}}c|}{\textbf{HD95$_{\smash{\scalebox{0.80}[1]{$\scriptscriptstyle std$}}}$ (mm)$\downarrow$}} 
& \multicolumn{1}{>{\columncolor{cyan!8}}c|}{\textbf{LR$_{\smash{\scalebox{0.80}[1]{$\scriptscriptstyle std$}}}$$\uparrow$}} 
& \multicolumn{3}{>{\columncolor{violet!6}}c}{\textbf{LLS$_{\smash{\scalebox{0.80}[1]{$\scriptscriptstyle std$}}}$ (\%)$\uparrow$}} \\
\cmidrule(lr{0.8em}){2-4}\cmidrule(lr{0.8em}){5-7}\cmidrule(lr{0.8em}){8-8}\cmidrule(lr{0.8em}){9-11}
& \kicon & \licon & \picon & \kicon & \licon & \picon & Avg & \kicon & \licon & \picon \\
 
\hline
\rowcolor{gray!10}\multicolumn{11}{l}{\textbf{Multimodal Foundation Models}}\\
\hline

M3D-LaMed$_{\smash{\scalebox{0.80}[1]{$\scriptscriptstyle 100\%$}}}$~\cite{2024M3D} & 35.3$_{\smash{\scalebox{0.80}[1]{$\scriptscriptstyle 32.7$}}}$ & 13.6$_{\smash{\scalebox{0.80}[1]{$\scriptscriptstyle 26.7$}}}$ & 19.9$_{\smash{\scalebox{0.80}[1]{$\scriptscriptstyle 26.1$}}}$ & 95.9$_{\smash{\scalebox{0.80}[1]{$\scriptscriptstyle 66.2$}}}$ & 103$_{\smash{\scalebox{0.80}[1]{$\scriptscriptstyle 45.9$}}}$ & 42.1$_{\smash{\scalebox{0.80}[1]{$\scriptscriptstyle 32.8$}}}$ & 34.6$_{\smash{\scalebox{0.80}[1]{$\scriptscriptstyle 41.6$}}}$ & 10.4$_{\smash{\scalebox{0.80}[1]{$\scriptscriptstyle 14.1$}}}$ & 2.62$_{\smash{\scalebox{0.80}[1]{$\scriptscriptstyle 5.94$}}}$ & 10.6$_{\smash{\scalebox{0.80}[1]{$\scriptscriptstyle 15.6$}}}$ \\

Merlin$_{\smash{\scalebox{0.80}[1]{$\scriptscriptstyle 100\%$}}}$~\cite{2024merlin} & 47.7$_{\smash{\scalebox{0.80}[1]{$\scriptscriptstyle 34.7$}}}$ & 27.5$_{\smash{\scalebox{0.80}[1]{$\scriptscriptstyle 30.8$}}}$ & 18.4$_{\smash{\scalebox{0.80}[1]{$\scriptscriptstyle 20.9$}}}$ & 130$_{\smash{\scalebox{0.80}[1]{$\scriptscriptstyle 128$}}}$ & 121$_{\smash{\scalebox{0.80}[1]{$\scriptscriptstyle 87.6$}}}$ & 70.0$_{\smash{\scalebox{0.80}[1]{$\scriptscriptstyle 36.4$}}}$ & 62.9$_{\smash{\scalebox{0.80}[1]{$\scriptscriptstyle 41.1$}}}$ & 21.7$_{\smash{\scalebox{0.80}[1]{$\scriptscriptstyle 22.1$}}}$ & 19.6$_{\smash{\scalebox{0.80}[1]{$\scriptscriptstyle 18.4$}}}$ & 5.77$_{\smash{\scalebox{0.80}[1]{$\scriptscriptstyle 9.99$}}}$ \\

SAT$_{\smash{\scalebox{0.80}[1]{$\scriptscriptstyle 100\%$}}}$~\cite{2025SAT} & 51.6$_{\smash{\scalebox{0.80}[1]{$\scriptscriptstyle 26.4$}}}$ & 34.9$_{\smash{\scalebox{0.80}[1]{$\scriptscriptstyle 21.3$}}}$ & 17.6$_{\smash{\scalebox{0.80}[1]{$\scriptscriptstyle 25.5$}}}$ & 96.7$_{\smash{\scalebox{0.80}[1]{$\scriptscriptstyle 76.2$}}}$ & 82.3$_{\smash{\scalebox{0.80}[1]{$\scriptscriptstyle 84.7$}}}$ & 54.9$_{\smash{\scalebox{0.80}[1]{$\scriptscriptstyle 68.1$}}}$ & 57.9$_{\smash{\scalebox{0.80}[1]{$\scriptscriptstyle 33.9$}}}$ & 23.7$_{\smash{\scalebox{0.80}[1]{$\scriptscriptstyle 15.7$}}}$ & 28.9$_{\smash{\scalebox{0.80}[1]{$\scriptscriptstyle 21.3$}}}$ & 11.3$_{\smash{\scalebox{0.80}[1]{$\scriptscriptstyle 10.5$}}}$ \\

R1Seg-3D$_{\smash{\scalebox{0.80}[1]{$\scriptscriptstyle 100\%$}}}$~\cite{2026R1SEG3D} & 28.9$_{\smash{\scalebox{0.80}[1]{$\scriptscriptstyle 29.7$}}}$ & 15.6$_{\smash{\scalebox{0.80}[1]{$\scriptscriptstyle 22.6$}}}$ & 6.47$_{\smash{\scalebox{0.80}[1]{$\scriptscriptstyle 15.0$}}}$ & 276$_{\smash{\scalebox{0.80}[1]{$\scriptscriptstyle 264$}}}$ & 196$_{\smash{\scalebox{0.80}[1]{$\scriptscriptstyle 177$}}}$ & 237$_{\smash{\scalebox{0.80}[1]{$\scriptscriptstyle 146$}}}$ & 43.7$_{\smash{\scalebox{0.80}[1]{$\scriptscriptstyle 39.6$}}}$ & 12.5$_{\smash{\scalebox{0.80}[1]{$\scriptscriptstyle 18.1$}}}$ & 15.3$_{\smash{\scalebox{0.80}[1]{$\scriptscriptstyle 17.8$}}}$ & 2.30$_{\smash{\scalebox{0.80}[1]{$\scriptscriptstyle 4.83$}}}$ \\

CT-CLIP$_{\smash{\scalebox{0.80}[1]{$\scriptscriptstyle 100\%$}}}$~\cite{2026ctclip} & 48.9$_{\smash{\scalebox{0.80}[1]{$\scriptscriptstyle 28.1$}}}$ & 28.6$_{\smash{\scalebox{0.80}[1]{$\scriptscriptstyle 24.5$}}}$ & 16.4$_{\smash{\scalebox{0.80}[1]{$\scriptscriptstyle 19.8$}}}$ & 127$_{\smash{\scalebox{0.80}[1]{$\scriptscriptstyle 86.8$}}}$ & 93.7$_{\smash{\scalebox{0.80}[1]{$\scriptscriptstyle 73.1$}}}$ & 81.6$_{\smash{\scalebox{0.80}[1]{$\scriptscriptstyle 56.4$}}}$ & 48.7$_{\smash{\scalebox{0.80}[1]{$\scriptscriptstyle 36.8$}}}$ & 18.9$_{\smash{\scalebox{0.80}[1]{$\scriptscriptstyle 17.3$}}}$ & 23.2$_{\smash{\scalebox{0.80}[1]{$\scriptscriptstyle 20.4$}}}$ & 9.38$_{\smash{\scalebox{0.80}[1]{$\scriptscriptstyle 11.5$}}}$ \\

\hline
\rowcolor{gray!10}\multicolumn{11}{l}{\textbf{Universal Segmentation Methods}}\\
\hline

UNet$_{\smash{\scalebox{0.80}[1]{$\scriptscriptstyle 100\%$}}}$~\cite{2015Unet} & 51.3$_{\smash{\scalebox{0.80}[1]{$\scriptscriptstyle 30.9$}}}$ & 31.7$_{\smash{\scalebox{0.80}[1]{$\scriptscriptstyle 24.6$}}}$ & 12.7$_{\smash{\scalebox{0.80}[1]{$\scriptscriptstyle 16.3$}}}$ & 180$_{\smash{\scalebox{0.80}[1]{$\scriptscriptstyle 152$}}}$ & 234$_{\smash{\scalebox{0.80}[1]{$\scriptscriptstyle 164$}}}$ & 105$_{\smash{\scalebox{0.80}[1]{$\scriptscriptstyle 151$}}}$ & 66.2$_{\smash{\scalebox{0.80}[1]{$\scriptscriptstyle 38.7$}}}$ & 30.2$_{\smash{\scalebox{0.80}[1]{$\scriptscriptstyle 25.4$}}}$ & 36.3$_{\smash{\scalebox{0.80}[1]{$\scriptscriptstyle 22.2$}}}$ & 6.83$_{\smash{\scalebox{0.80}[1]{$\scriptscriptstyle 10.8$}}}$ \\

nnUNetV2$_{\smash{\scalebox{0.80}[1]{$\scriptscriptstyle 100\%$}}}$~\cite{2021nnunet}& 55.7$_{\smash{\scalebox{0.80}[1]{$\scriptscriptstyle 34.2$}}}$ & 34.8$_{\smash{\scalebox{0.80}[1]{$\scriptscriptstyle 28.1$}}}$ & 21.1$_{\smash{\scalebox{0.80}[1]{$\scriptscriptstyle 22.3$}}}$ & 152$_{\smash{\scalebox{0.80}[1]{$\scriptscriptstyle 129$}}}$ & 72.7$_{\smash{\scalebox{0.80}[1]{$\scriptscriptstyle 65.1$}}}$ & 109$_{\smash{\scalebox{0.80}[1]{$\scriptscriptstyle 85.4$}}}$ & 71.4$_{\smash{\scalebox{0.80}[1]{$\scriptscriptstyle 40.2$}}}$ & 42.2$_{\smash{\scalebox{0.80}[1]{$\scriptscriptstyle 29.0$}}}$ & 27.2$_{\smash{\scalebox{0.80}[1]{$\scriptscriptstyle 26.3$}}}$ & 11.6$_{\smash{\scalebox{0.80}[1]{$\scriptscriptstyle 12.3$}}}$ \\

STUNet$_{\smash{\scalebox{0.80}[1]{$\scriptscriptstyle 100\%$}}}~\cite{Huang2023STUNetSA}$ & 53.0$_{\smash{\scalebox{0.80}[1]{$\scriptscriptstyle 35.3$}}}$ & 30.8$_{\smash{\scalebox{0.80}[1]{$\scriptscriptstyle 25.6$}}}$ & \second{22.6}$_{\smash{\scalebox{0.80}[1]{$\scriptscriptstyle 23.2$}}}$ & 136$_{\smash{\scalebox{0.80}[1]{$\scriptscriptstyle 112$}}}$ & 119$_{\smash{\scalebox{0.80}[1]{$\scriptscriptstyle 104$}}}$ & 111$_{\smash{\scalebox{0.80}[1]{$\scriptscriptstyle 101$}}}$ & 65.5$_{\smash{\scalebox{0.80}[1]{$\scriptscriptstyle 40.2$}}}$ & 33.8$_{\smash{\scalebox{0.80}[1]{$\scriptscriptstyle 28.3$}}}$ & 33.2$_{\smash{\scalebox{0.80}[1]{$\scriptscriptstyle 26.5$}}}$ & 12.8$_{\smash{\scalebox{0.80}[1]{$\scriptscriptstyle 15.6$}}}$ \\

Swin{\scriptsize UNETR}$_{\smash{\scalebox{0.80}[1]{$\scriptscriptstyle 100\%$}}}$~\cite{2022swin} & 56.5$_{\smash{\scalebox{0.80}[1]{$\scriptscriptstyle 32.2$}}}$ & 34.2$_{\smash{\scalebox{0.80}[1]{$\scriptscriptstyle 30.2$}}}$ & 22.2$_{\smash{\scalebox{0.80}[1]{$\scriptscriptstyle 19.5$}}}$ & 130$_{\smash{\scalebox{0.80}[1]{$\scriptscriptstyle 115$}}}$ & 153$_{\smash{\scalebox{0.80}[1]{$\scriptscriptstyle 133$}}}$ & 105$_{\smash{\scalebox{0.80}[1]{$\scriptscriptstyle 79.9$}}}$ & 71.4$_{\smash{\scalebox{0.80}[1]{$\scriptscriptstyle 37.5$}}}$ & 36.6$_{\smash{\scalebox{0.80}[1]{$\scriptscriptstyle 26.3$}}}$ & {31.7}$_{\smash{\scalebox{0.80}[1]{$\scriptscriptstyle 25.6$}}}$ & 13.4$_{\smash{\scalebox{0.80}[1]{$\scriptscriptstyle 13.5$}}}$ \\

Medformer$_{\smash{\scalebox{0.80}[1]{$\scriptscriptstyle 100\%$}}}$~\cite{medformer2023} & 59.8$_{\smash{\scalebox{0.80}[1]{$\scriptscriptstyle 28.0$}}}$ & 44.3$_{\smash{\scalebox{0.80}[1]{$\scriptscriptstyle 25.2$}}}$ & 16.4$_{\smash{\scalebox{0.80}[1]{$\scriptscriptstyle 22.3$}}}$ & \second{71.0}$_{\smash{\scalebox{0.80}[1]{$\scriptscriptstyle 149$}}}$ & 90.1$_{\smash{\scalebox{0.80}[1]{$\scriptscriptstyle 127$}}}$ & 36.5$_{\smash{\scalebox{0.80}[1]{$\scriptscriptstyle 30.2$}}}$ & 73.2$_{\smash{\scalebox{0.80}[1]{$\scriptscriptstyle 39.0$}}}$ & 39.2$_{\smash{\scalebox{0.80}[1]{$\scriptscriptstyle 22.6$}}}$ & {35.3}$_{\smash{\scalebox{0.80}[1]{$\scriptscriptstyle 27.8$}}}$ & \second{15.9}$_{\smash{\scalebox{0.80}[1]{$\scriptscriptstyle 17.0$}}}$ \\

Med-SAM3$_{\smash{\scalebox{0.80}[1]{$\scriptscriptstyle 100\%$}}}$~\cite{2025medsam3} & 34.5$_{\smash{\scalebox{0.80}[1]{$\scriptscriptstyle 25.4$}}}$ & 28.1$_{\smash{\scalebox{0.80}[1]{$\scriptscriptstyle 28.9$}}}$ & 15.1$_{\smash{\scalebox{0.80}[1]{$\scriptscriptstyle 17.9$}}}$ & 141$_{\smash{\scalebox{0.80}[1]{$\scriptscriptstyle 71.5$}}}$ & 120$_{\smash{\scalebox{0.80}[1]{$\scriptscriptstyle 67.8$}}}$ & 127$_{\smash{\scalebox{0.80}[1]{$\scriptscriptstyle 79.3$}}}$ & 68.5$_{\smash{\scalebox{0.80}[1]{$\scriptscriptstyle 40.1$}}}$ & 23.3$_{\smash{\scalebox{0.80}[1]{$\scriptscriptstyle 19.6$}}}$ & 25.9$_{\smash{\scalebox{0.80}[1]{$\scriptscriptstyle 23.6$}}}$ & 11.8$_{\smash{\scalebox{0.80}[1]{$\scriptscriptstyle 13.3$}}}$ \\
\hline
\rowcolor{gray!10}\multicolumn{11}{l}{\textbf{Advanced Methods}}\\
\hline

R-Super$_{\smash{\scalebox{0.80}[1]{$\scriptscriptstyle 10\%$}}}$~\cite{bassi2025learning} & 53.3$_{\smash{\scalebox{0.80}[1]{$\scriptscriptstyle 32.4$}}}$ & 37.5$_{\smash{\scalebox{0.80}[1]{$\scriptscriptstyle 27.8$}}}$ & 15.1$_{\smash{\scalebox{0.80}[1]{$\scriptscriptstyle 19.6$}}}$ & 162$_{\smash{\scalebox{0.80}[1]{$\scriptscriptstyle 193$}}}$ & 128$_{\smash{\scalebox{0.80}[1]{$\scriptscriptstyle 139$}}}$ & 99.6$_{\smash{\scalebox{0.80}[1]{$\scriptscriptstyle 85.3$}}}$ & 73.9$_{\smash{\scalebox{0.80}[1]{$\scriptscriptstyle 36.9$}}}$ & 32.9$_{\smash{\scalebox{0.80}[1]{$\scriptscriptstyle 25.5$}}}$ & 33.5$_{\smash{\scalebox{0.80}[1]{$\scriptscriptstyle 19.8$}}}$ & 6.71$_{\smash{\scalebox{0.80}[1]{$\scriptscriptstyle 12.5$}}}$ \\

R-Super$_{\smash{\scalebox{0.80}[1]{$\scriptscriptstyle 25\%$}}}$~\cite{bassi2025learning} & 54.3$_{\smash{\scalebox{0.80}[1]{$\scriptscriptstyle 32.0$}}}$ & 39.6$_{\smash{\scalebox{0.80}[1]{$\scriptscriptstyle 31.3$}}}$ & 20.5$_{\smash{\scalebox{0.80}[1]{$\scriptscriptstyle 25.3$}}}$ & 101$_{\smash{\scalebox{0.80}[1]{$\scriptscriptstyle 128$}}}$ & 74.4$_{\smash{\scalebox{0.80}[1]{$\scriptscriptstyle 73.5$}}}$ & 85.1$_{\smash{\scalebox{0.80}[1]{$\scriptscriptstyle 73.2$}}}$ & 74.4$_{\smash{\scalebox{0.80}[1]{$\scriptscriptstyle 36.4$}}}$ & 37.2$_{\smash{\scalebox{0.80}[1]{$\scriptscriptstyle 26.8$}}}$ & 37.1$_{\smash{\scalebox{0.80}[1]{$\scriptscriptstyle 23.6$}}}$ & 15.3$_{\smash{\scalebox{0.80}[1]{$\scriptscriptstyle 14.8$}}}$ \\

R-Super$_{\smash{\scalebox{0.80}[1]{$\scriptscriptstyle 100\%$}}}$~\cite{bassi2025learning} & \second{63.9}$_{\smash{\scalebox{0.80}[1]{$\scriptscriptstyle 28.5$}}}$ & \best{49.2}$_{\smash{\scalebox{0.80}[1]{$\scriptscriptstyle 26.9$}}}$ & \second{22.6}$_{\smash{\scalebox{0.80}[1]{$\scriptscriptstyle 25.5$}}}$ & 81.1$_{\smash{\scalebox{0.80}[1]{$\scriptscriptstyle 118$}}}$ & \second{66.6}$_{\smash{\scalebox{0.80}[1]{$\scriptscriptstyle 53.2$}}}$ & \second{33.4}$_{\smash{\scalebox{0.80}[1]{$\scriptscriptstyle 26.9$}}}$ & 74.6$_{\smash{\scalebox{0.80}[1]{$\scriptscriptstyle 38.3$}}}$ & \second{42.9}$_{\smash{\scalebox{0.80}[1]{$\scriptscriptstyle 23.7$}}}$ & \best{40.9}$_{\smash{\scalebox{0.80}[1]{$\scriptscriptstyle 28.7$}}}$ & 15.5$_{\smash{\scalebox{0.80}[1]{$\scriptscriptstyle 17.4$}}}$ \\

\textbf{GLeVE (Ours)}$_{\smash{\scalebox{0.80}[1]{$\scriptscriptstyle 10\%$}}}$ 
& 53.9$_{\smash{\scalebox{0.80}[1]{$\scriptscriptstyle 32.1$}}}$ 
& 38.2$_{\smash{\scalebox{0.80}[1]{$\scriptscriptstyle 29.4$}}}$ 
& 18.4$_{\smash{\scalebox{0.80}[1]{$\scriptscriptstyle 23.0$}}}$ 
& 113$_{\smash{\scalebox{0.80}[1]{$\scriptscriptstyle 148$}}}$ 
& 96.1$_{\smash{\scalebox{0.80}[1]{$\scriptscriptstyle 131$}}}$ 
& 62.7$_{\smash{\scalebox{0.80}[1]{$\scriptscriptstyle 92.4$}}}$ 
& 74.1$_{\smash{\scalebox{0.80}[1]{$\scriptscriptstyle 40.1$}}}$ 
& 35.4$_{\smash{\scalebox{0.80}[1]{$\scriptscriptstyle 25.1$}}}$ 
& 33.9$_{\smash{\scalebox{0.80}[1]{$\scriptscriptstyle 23.7$}}}$ 
& 12.8$_{\smash{\scalebox{0.80}[1]{$\scriptscriptstyle 18.4$}}}$ \\

\textbf{GLeVE (Ours)}$_{\smash{\scalebox{0.80}[1]{$\scriptscriptstyle 25\%$}}}$ 
& 56.6$_{\smash{\scalebox{0.80}[1]{$\scriptscriptstyle 29.8$}}}$ 
& 42.9$_{\smash{\scalebox{0.80}[1]{$\scriptscriptstyle 27.1$}}}$ 
& 19.3$_{\smash{\scalebox{0.80}[1]{$\scriptscriptstyle 21.4$}}}$ 
& 98.4$_{\smash{\scalebox{0.80}[1]{$\scriptscriptstyle 132$}}}$ 
& 83.3$_{\smash{\scalebox{0.80}[1]{$\scriptscriptstyle 118$}}}$ 
& 54.6$_{\smash{\scalebox{0.80}[1]{$\scriptscriptstyle 85.7$}}}$ 
& \second{74.9}$_{\smash{\scalebox{0.80}[1]{$\scriptscriptstyle 37.8$}}}$ 
& 38.1$_{\smash{\scalebox{0.80}[1]{$\scriptscriptstyle 24.5$}}}$ 
& 37.7$_{\smash{\scalebox{0.80}[1]{$\scriptscriptstyle 21.4$}}}$ 
& 15.5$_{\smash{\scalebox{0.80}[1]{$\scriptscriptstyle 18.6$}}}$ \\

\textbf{GLeVE (Ours)}$_{\smash{\scalebox{0.80}[1]{$\scriptscriptstyle 100\%$}}}$ 
& \best{65.8}$_{\smash{\scalebox{0.80}[1]{$\scriptscriptstyle 26.2$}}}$ 
& \second{48.7}$_{\smash{\scalebox{0.80}[1]{$\scriptscriptstyle 23.9$}}}$ 
& \best{27.8}$_{\smash{\scalebox{0.80}[1]{$\scriptscriptstyle 19.6$}}}$ 
& \best{69.5}$_{\smash{\scalebox{0.80}[1]{$\scriptscriptstyle 109$}}}$ 
& \best{60.2}$_{\smash{\scalebox{0.80}[1]{$\scriptscriptstyle 96.4$}}}$ 
& \best{29.5}$_{\smash{\scalebox{0.80}[1]{$\scriptscriptstyle 64.8$}}}$ 
& \best{76.2}$_{\smash{\scalebox{0.80}[1]{$\scriptscriptstyle 35.3$}}}$ 
& \best{44.3}$_{\smash{\scalebox{0.80}[1]{$\scriptscriptstyle 21.2$}}}$ 
& \second{38.9}$_{\smash{\scalebox{0.80}[1]{$\scriptscriptstyle 20.4$}}}$ 
& \best{17.8}$_{\smash{\scalebox{0.80}[1]{$\scriptscriptstyle 14.8$}}}$ \\

\bottomrule
\end{tabular*}%
}
\end{table}

\begin{figure}[t!]
\includegraphics[width=\textwidth]{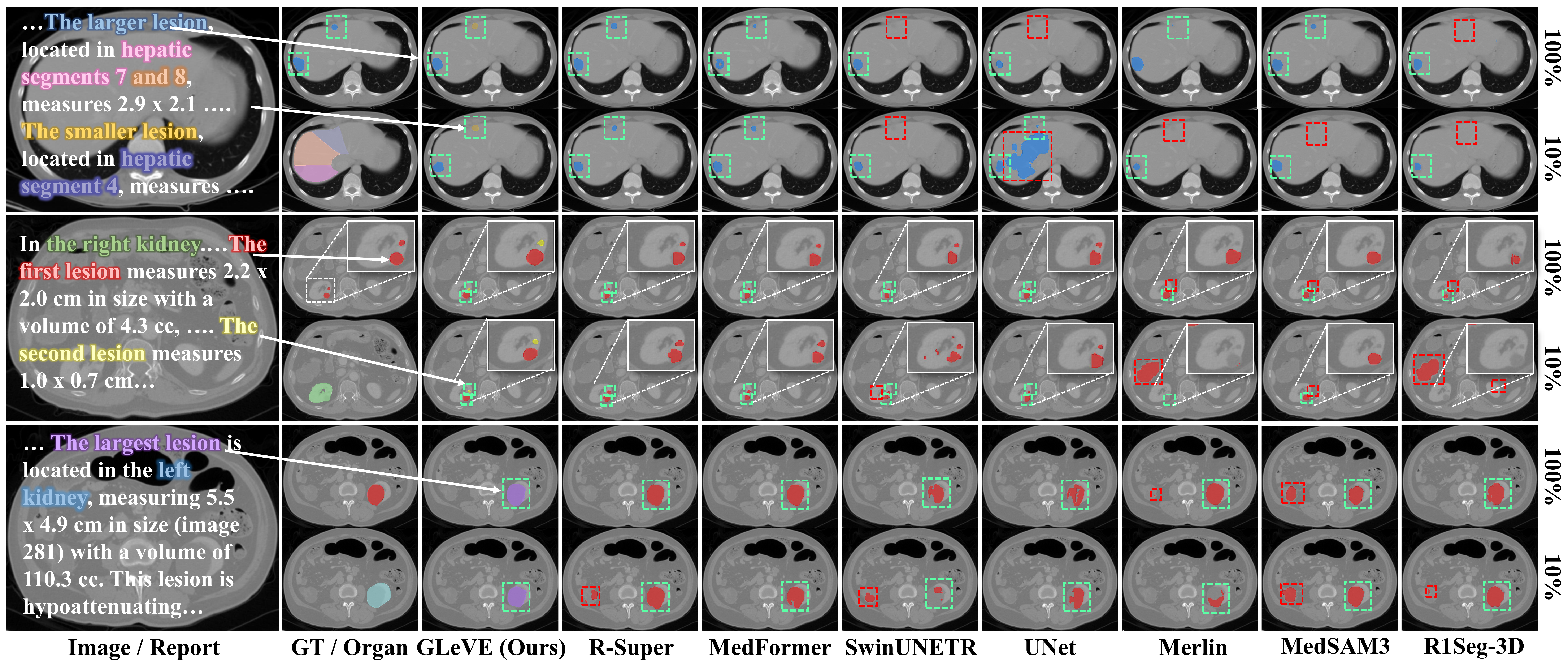}
\caption{Qualitative comparison of grounding results under 10\% and 100\% mask supervision, highlighting localization accuracy and multi-lesion disambiguation.}
\label{Com}
\end{figure}

\subsection{Experiment Results}

\textbf{Comparative Experiments.}
As shown in Table~\ref{tab-com} and Fig.~\ref{Com}, with full mask annotations and report supervision, \textbf{GLeVE} consistently surpasses multimodal foundation models (e.g., CT-CLIP~\cite{2026ctclip}), universal segmentation methods (e.g., MedFormer~\cite{medformer2023}), and an advanced report-assisted method R-Super~\cite{bassi2025learning}. Averaged across organs, it increases Dice by 7.26\% over the best universal segmentation method and reduces HD95 by 7.3~mm relative to the advanced method, indicating improved global geometric consistency and fewer extreme boundary errors. Lesion-level performance is likewise strong, reaching 76.2\% LR and 33.7\% LLS. Under weak supervision with only 10\% or 25\% masks, \textbf{GLeVE} remains competitive and often outperforms most alternatives, highlighting robustness to annotation scarcity and improved data efficiency.

\begin{table}[t!]
\caption{Ablation analysis of key components under 100\% mask supervision.}
\label{tab-abl}
\centering
\fontsize{8pt}{10pt}\selectfont
\setlength{\tabcolsep}{0.65pt}
{
\renewcommand{\arraystretch}{0.88}
\begin{tabular*}{\linewidth}{@{\extracolsep{\fill}}l|ccc|ccc|c|ccc}
\toprule
\multirow{2.3}{*}{\textbf{Method}} 
& \multicolumn{3}{>{\columncolor{green!6}}c|}{\textbf{Dice$_{\smash{\scalebox{0.80}[1]{$\scriptscriptstyle std$}}}$ (\%)$\uparrow$}} 
& \multicolumn{3}{>{\columncolor{orange!5}}c|}{\textbf{HD95$_{\smash{\scalebox{0.80}[1]{$\scriptscriptstyle std$}}}$ (mm)$\downarrow$}} 
& \multicolumn{1}{>{\columncolor{cyan!8}}c|}{\textbf{LR$_{\smash{\scalebox{0.80}[1]{$\scriptscriptstyle std$}}}$$\uparrow$}} 
& \multicolumn{3}{>{\columncolor{violet!6}}c}{\textbf{LLS$_{\smash{\scalebox{0.80}[1]{$\scriptscriptstyle std$}}}$ (\%)$\uparrow$}} \\
\cmidrule(lr{0.8em}){2-4}\cmidrule(lr{0.8em}){5-7}\cmidrule(lr{0.8em}){8-8}\cmidrule(lr{0.8em}){9-11}
& \kicon & \licon & \picon & \kicon & \licon & \picon & Avg & \kicon & \licon & \picon \\
    \hline
    
w/o LeQu 
& 59.5$_{\smash{\scalebox{0.80}[1]{$\scriptscriptstyle 28.4$}}}$ 
& 42.5$_{\smash{\scalebox{0.80}[1]{$\scriptscriptstyle 25.7$}}}$ 
& 23.5$_{\smash{\scalebox{0.80}[1]{$\scriptscriptstyle 20.9$}}}$ 
& 84.6$_{\smash{\scalebox{0.80}[1]{$\scriptscriptstyle 124$}}}$ 
& 72.0$_{\smash{\scalebox{0.80}[1]{$\scriptscriptstyle 108$}}}$ 
& 39.3$_{\smash{\scalebox{0.80}[1]{$\scriptscriptstyle 76.5$}}}$ 
& 73.3$_{\smash{\scalebox{0.80}[1]{$\scriptscriptstyle 37.2$}}}$ 
& 40.9$_{\smash{\scalebox{0.80}[1]{$\scriptscriptstyle 22.1$}}}$ 
& 32.8$_{\smash{\scalebox{0.80}[1]{$\scriptscriptstyle 20.7$}}}$ 
& 16.4$_{\smash{\scalebox{0.80}[1]{$\scriptscriptstyle 16.2$}}}$ \\

w/o AnVer 
& 55.9$_{\smash{\scalebox{0.80}[1]{$\scriptscriptstyle 31.6$}}}$ 
& 39.7$_{\smash{\scalebox{0.80}[1]{$\scriptscriptstyle 28.9$}}}$ 
& 20.3$_{\smash{\scalebox{0.80}[1]{$\scriptscriptstyle 23.1$}}}$ 
& 89.8$_{\smash{\scalebox{0.80}[1]{$\scriptscriptstyle 134$}}}$ 
& 75.1$_{\smash{\scalebox{0.80}[1]{$\scriptscriptstyle 121$}}}$ 
& 47.6$_{\smash{\scalebox{0.80}[1]{$\scriptscriptstyle 84.9$}}}$ 
& 69.3$_{\smash{\scalebox{0.80}[1]{$\scriptscriptstyle 40.5$}}}$ 
& 34.6$_{\smash{\scalebox{0.80}[1]{$\scriptscriptstyle 24.6$}}}$ 
& 30.5$_{\smash{\scalebox{0.80}[1]{$\scriptscriptstyle 22.9$}}}$ 
& 13.9$_{\smash{\scalebox{0.80}[1]{$\scriptscriptstyle 17.8$}}}$ \\

w/o OcRe 
& \second{63.7}$_{\smash{\scalebox{0.80}[1]{$\scriptscriptstyle 26.9$}}}$ 
& \second{44.3}$_{\smash{\scalebox{0.80}[1]{$\scriptscriptstyle 24.1$}}}$ 
& \second{25.4}$_{\smash{\scalebox{0.80}[1]{$\scriptscriptstyle 19.8$}}}$ 
& \second{74.8}$_{\smash{\scalebox{0.80}[1]{$\scriptscriptstyle 112$}}}$ 
& \second{64.0}$_{\smash{\scalebox{0.80}[1]{$\scriptscriptstyle 99.6$}}}$ 
& \second{33.0}$_{\smash{\scalebox{0.80}[1]{$\scriptscriptstyle 69.2$}}}$ 
& \second{75.1}$_{\smash{\scalebox{0.80}[1]{$\scriptscriptstyle 36.6$}}}$ 
& \second{42.2}$_{\smash{\scalebox{0.80}[1]{$\scriptscriptstyle 21.7$}}}$ 
& \second{33.9}$_{\smash{\scalebox{0.80}[1]{$\scriptscriptstyle 19.8$}}}$ 
& \second{17.6}$_{\smash{\scalebox{0.80}[1]{$\scriptscriptstyle 15.1$}}}$ \\
    
\textbf{GLeVE (Ours)}
& \best{65.8}$_{\smash{\scalebox{0.80}[1]{$\scriptscriptstyle 26.2$}}}$ 
& \best{48.7}$_{\smash{\scalebox{0.80}[1]{$\scriptscriptstyle 23.9$}}}$ 
& \best{27.8}$_{\smash{\scalebox{0.80}[1]{$\scriptscriptstyle 19.6$}}}$ 
& \best{69.5}$_{\smash{\scalebox{0.80}[1]{$\scriptscriptstyle 109$}}}$ 
& \best{60.2}$_{\smash{\scalebox{0.80}[1]{$\scriptscriptstyle 96.4$}}}$ 
& \best{29.5}$_{\smash{\scalebox{0.80}[1]{$\scriptscriptstyle 64.8$}}}$ 
& \best{76.2}$_{\smash{\scalebox{0.80}[1]{$\scriptscriptstyle 35.3$}}}$ 
& \best{44.3}$_{\smash{\scalebox{0.80}[1]{$\scriptscriptstyle 21.2$}}}$ 
& \best{38.9}$_{\smash{\scalebox{0.80}[1]{$\scriptscriptstyle 20.4$}}}$ 
& \best{17.8}$_{\smash{\scalebox{0.80}[1]{$\scriptscriptstyle 14.8$}}}$ \\

    \hline
    \end{tabular*}%
    }
    \end{table}

\begin{figure}[t!]
\includegraphics[width=\textwidth]{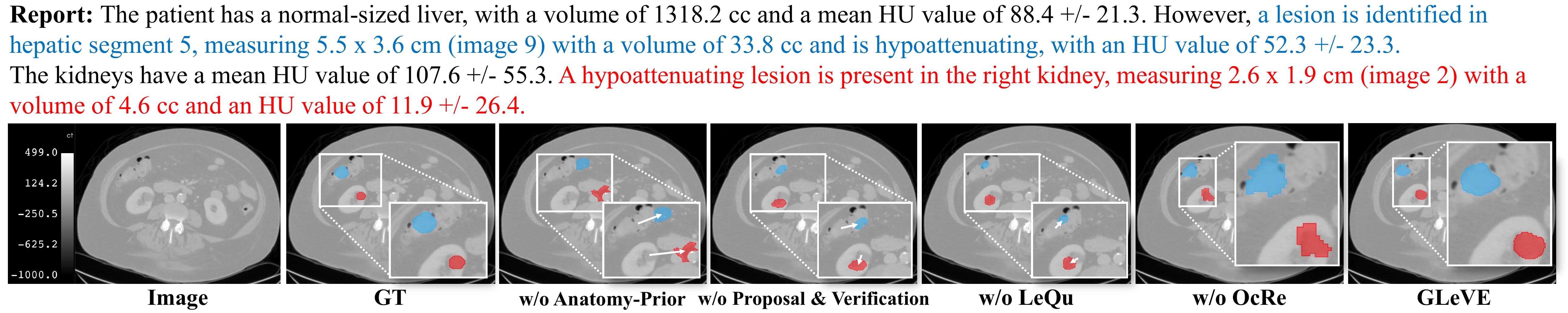}
\caption{Qualitative ablation results of GLeVE, showing the effect of removing Anatomical-Prior, Proposal \& Verification, LeQu, and OcRe on lesion grounding and boundary refinement. Arrows indicate localization center offsets.}
\label{Abl}
\end{figure}

\textbf{Ablation Study.}
As reported in Table~\ref{tab-abl} and Fig.~\ref{Abl}, replacing LeQu with a text encoder applied to structured descriptions reduces LR by 2.9\%, indicating that explicit relational modeling preserves lesion semantics and enhances representation discriminability. Disabling AnVer reduces the organ-averaged LLS by 7.3\%, indicating that candidate filtering and verification are critical for semantic alignment. Removing OcRe increases the organ-averaged HD95 by 4.2~mm, highlighting the contribution of hierarchical recursive refinement to accurate boundary delineation. Overall, the three modules are complementary in semantic modeling, verification, and refinement, collectively enabling robust localization.

\section{Conclusion}

In this study, we propose \textbf{GLeVE} for verifiable, pixel-accurate lesion grounding from radiology reports in 3D CT. By modeling each lesion description as an atomic unit, \textbf{GLeVE} learns relation-aware lesion queries, enforces anatomy-consistent one-to-one text-lesion matching via soft modulation and region-level verification, and refines masks with octree-based autoregressive coarse-to-fine decoding; a report-driven objective further enables training under missing masks. Experiments on AbdomenAtlas 3.0 demonstrate consistent improvements over advanced methods, supporting robust grounding in challenging multi-lesion settings. Future work will explore extending the framework to broader anatomical regions and more diverse disease categories.

\subsubsection{Acknowledgements}

This work was supported by the National Natural Science Foundation of China
(No. 62472133) and the Fundamental Research Funds for the Provincial
Universities of Zhejiang (No. GK259909299001-006).

\subsubsection{Disclosure of Interests} The authors declare no competing interests.

\bibliographystyle{splncs04}
\bibliography{Paper-2407}
\end{document}